\documentclass[10pt,twocolumn,letterpaper]{article}

\usepackage{iccv}
\usepackage{times}
\usepackage{epsfig}
\usepackage{graphicx}
\usepackage{amsmath}
\usepackage{amssymb}
\usepackage{booktabs}
\usepackage{bbding}
\usepackage{subcaption}
\usepackage{multirow}

\usepackage[pagebackref=true,breaklinks=true,letterpaper=true,colorlinks,bookmarks=false]{hyperref}

\iccvfinalcopy 

\def\iccvPaperID{6636} 

\ificcvfinal\fi

\begin{document}

\title{Neural 3D Scene Reconstruction from Multiple 2D Images

without 3D Supervision}

\author{Yi Guo\textsuperscript{\rm 1}, Che Sun\textsuperscript{\rm 1}, Yunde Jia\textsuperscript{\rm 2,\rm 1}, and Yuwei Wu\textsuperscript{\rm 1,\rm2}\\
\textsuperscript{\rm 1}Beijing Key Laboratory of Intellegent Information Technology, \\ School of Computer Science \& Technology, Beijing Institute of Technology, China \\
\textsuperscript{\rm 2}Guangdong Laboratory of Machine Perception and Intelligent Computing, \\
   Shenzhen MSU-BIT University, China \\
{\tt\small \{guoyi,sunche,jiayunde,,wuyuwei\}@bit.edu.cn}}

\maketitle
\ificcvfinal\thispagestyle{empty}\fi

\begin{abstract}
   Neural 3D scene reconstruction methods have achieved impressive performance when reconstructing complex geometry and low-textured regions in indoor scenes. 
   However, these methods heavily rely on 3D data which is costly and time-consuming to obtain in real world. In this paper, we propose a novel neural reconstruction method that reconstructs scenes using sparse depth under the plane constraints without 3D supervision. 
   We introduce a signed distance function field, a color field, and a probability field to represent a scene.
   We optimize these fields to reconstruct the scene by using differentiable ray marching with accessible 2D images as supervision.
   We improve the reconstruction quality of complex geometry scene regions with sparse depth obtained by using the geometric constraints.
   The geometric constraints project 3D points on the surface to similar-looking regions with similar features in different 2D images.
   We impose the plane constraints to make large planes parallel or vertical to the indoor floor. Both two constraints help reconstruct accurate and smooth geometry structures of the scene. Without 3D supervision, our method achieves competitive performance compared with existing methods that use 3D supervision on the ScanNet dataset.
\end{abstract}

\section{Introduction}
\label{sec:intro}

\begin{figure}
  \centering
  \subfloat[Ground Truth]{\label{fig1_a}\includegraphics[width=0.215\textwidth]{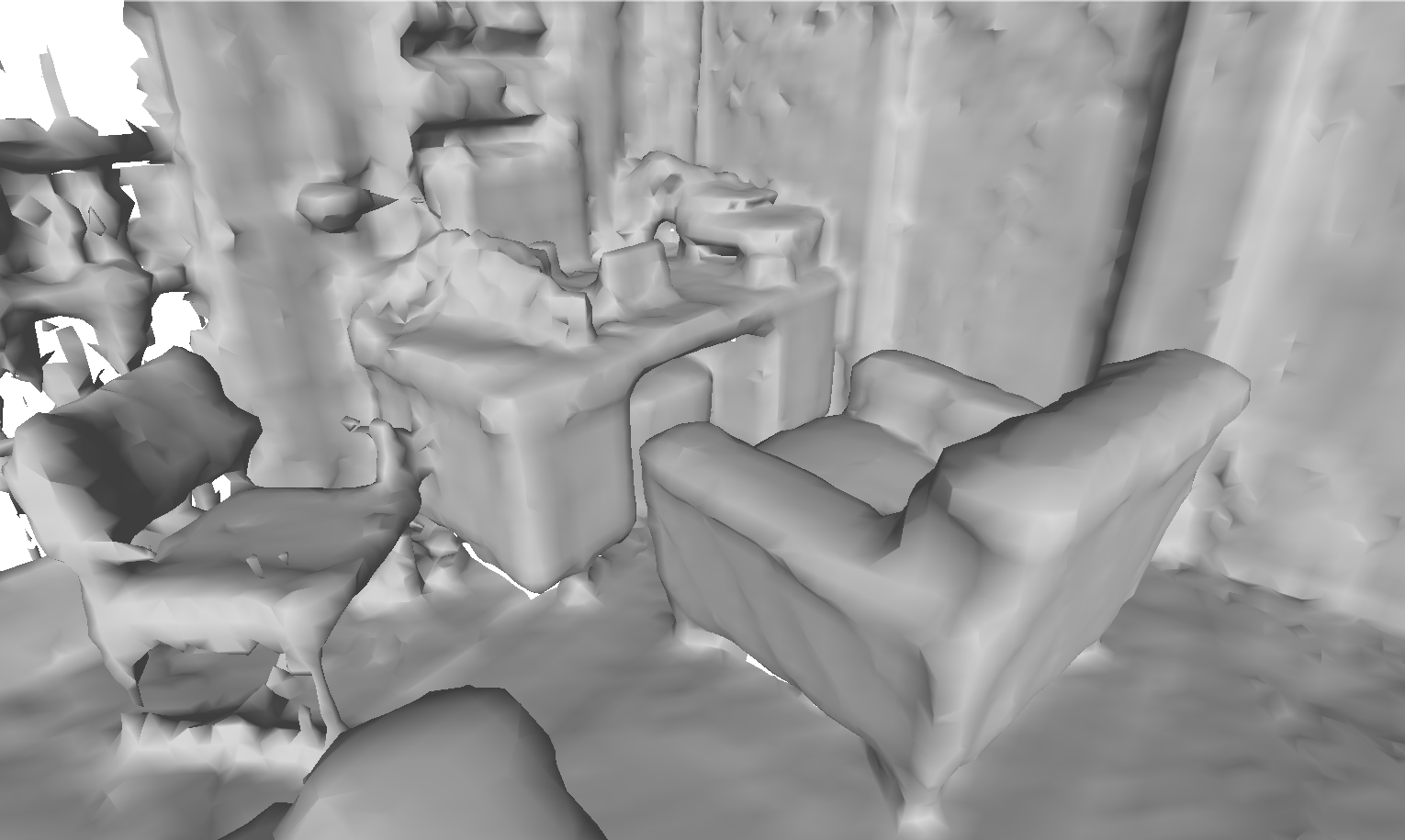}}
  \hspace{0.7mm}
  \subfloat[VolSDF \cite{yariv2021volume}]{\label{fig1_b}\includegraphics[width=0.215\textwidth]{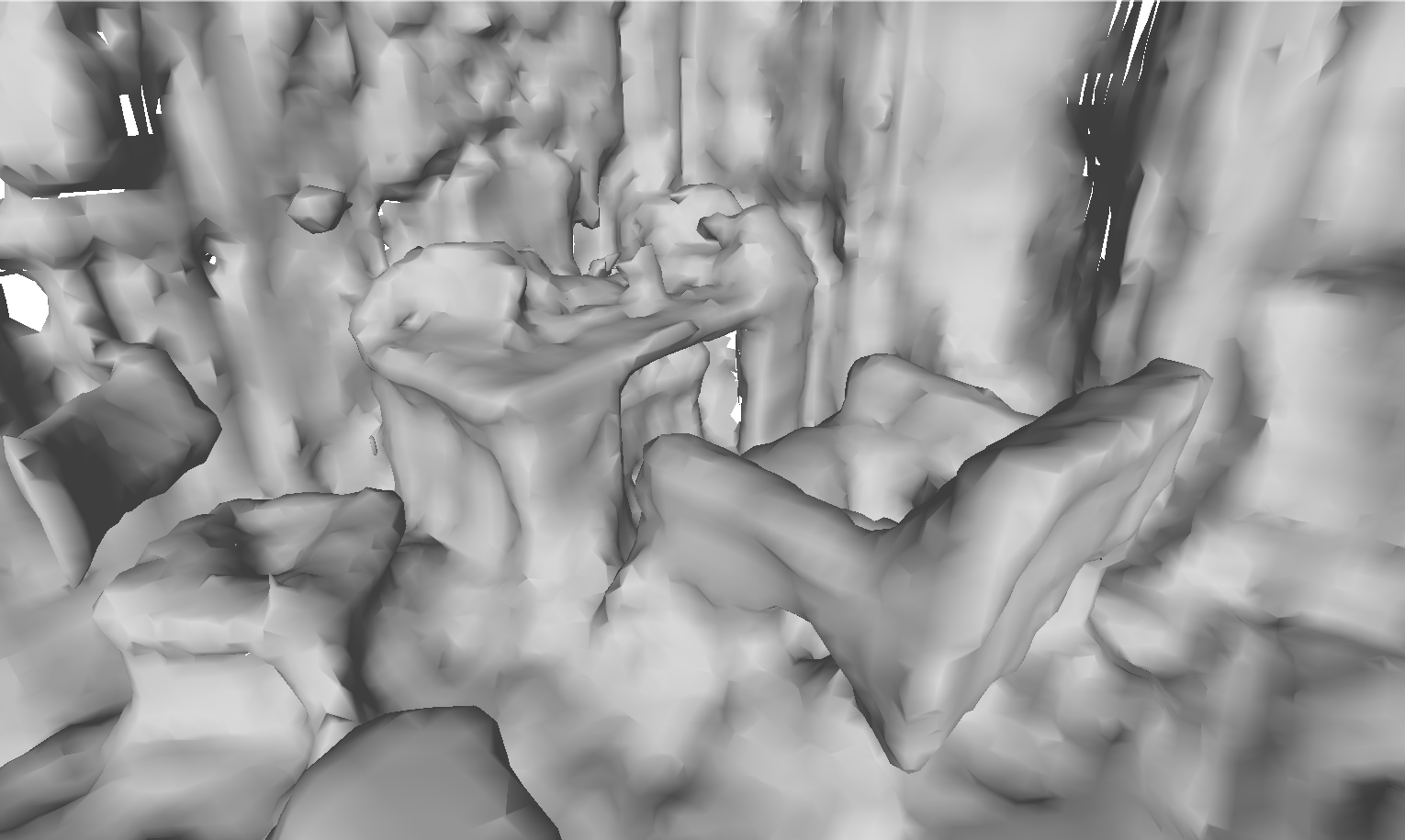}}
  
  \subfloat[Manhattan-SDF \cite{guo2022neural}]{\label{fig1_c}\includegraphics[width=0.215\textwidth]{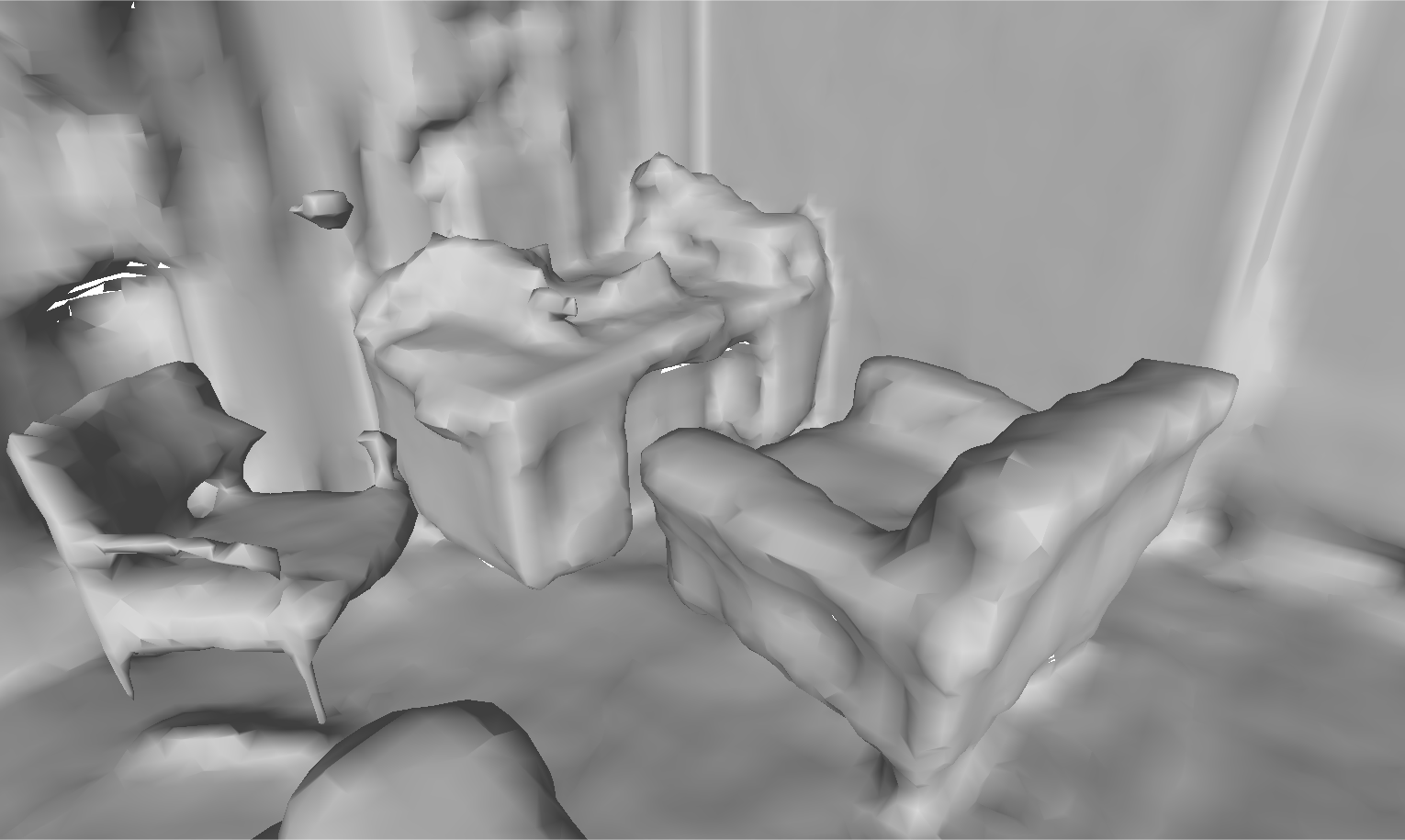}}
  \hspace{0.7mm}
  \subfloat[Ours]{\label{fig1_d}\includegraphics[width=0.215\textwidth]{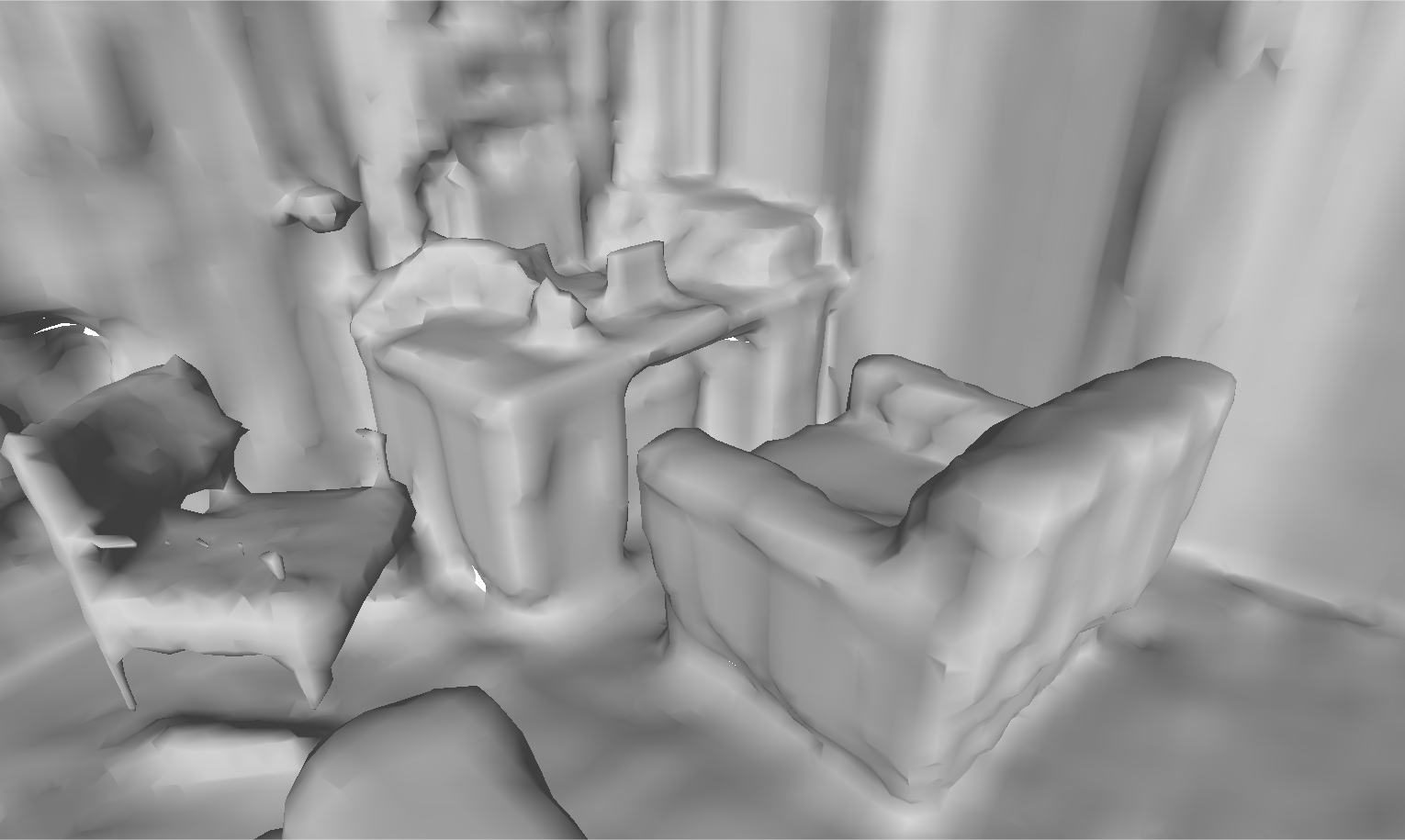}}
  \caption{\textbf{Comparisons of neural reconstructions.} (a) The ground truth of the scene. (b) The reconstruction result of VolSDF using 2D color supervision. (c) The reconstruction result of Manhattan-SDF with dense depth predictions and the Manhattan assumption. (d) Our result with 2D color supervision by using sparse depth and constraints.  
  }
  \label{fig1}
 \end{figure}

Learning-based 3D scene reconstruction methods have achieved good performance from multi-view images. 
Some works \cite{yao2018mvsnet, cheng2020deep, tang2018ba} use deep networks to perform some processes in conventional formulation, such as extracting features, matching features, estimating depth, and fusing depth.
Others \cite{murez2020atlas, sun2021neuralrecon} use deep networks to recover 3D representations directly from a whole sequence of images.
However, these learning-based methods heavily rely on 3D supervised data to optimize the large scale parameters in deep networks.
And obtaining 3D supervised data of real scenes is costly and time-consuming. 
In contrast, acquiring 2D image data is simple and inexpensive, which inspires researchers to use 2D images to reconstruct 3D scenes.

Several works \cite{wang2021neus, yariv2021volume, yariv2020multiview, oechsle2021unisurf} reconstruct the 3D geometry structures of scenes from 2D images based on NeRF \cite{mildenhall2020nerf}. 
These methods represent scenes as neural implicit fields and combine the surface representation with volume rendering, and achieve impressive results.
However, they do not work well in reconstructing some challenging indoor scenes.
For example, VoISDF \cite{yariv2021volume} fails to reconstruct the regions with complex geometric structures and large low-textured planes, as shown in Figure \ref{fig1_b}.

In order for handling this issue, Guo \textit{et al.} \cite{guo2022neural} use dense depth and the Manhattan-world assumption to reconstruct indoor scenes with 2D images, and achieve impressive performance. 
However, dense depth from 2D images is inevitably noisy, incurring poor reconstruction quality of the planes, such as tables and chairs shown in Figure \ref{fig1_c}, because these planes do not fit the Manhattan-world assumption.
To improve the work \cite{guo2022neural}, Wang \textit{et al.} \cite{wang2022neuris} and Yu \textit{et al.} \cite{yu2022monosdf} use pre-trained networks to obtain extra 3D data, such as dense depth and normal maps for improving the reconstruction quality of plane regions.
However, the pre-trained networks have to acquire extra 3D supervised data for fine-tuning.
Obtaining sufficient 3D supervised data is often costly and time-consuming.
Our work, without using any 3D supervised data, focuses on improving the reconstruction quality of indoor scenes.



In this paper, we propose a novel method that reconstructs indoor scenes using spare depth maps under the plane constraints from 2D images.
The sparse depth from 2D images is easily obtained and has less noise compared with dense depth.
The plane constraints would help reconstruct low-textured plane regions in the scene, not only walls and floors in the Manhattan-world assumption but also planes of tables and chairs, as shown in Figure \ref{fig1_d}.
That is to say, our plane constraints are more general, and have a wider scope of applications and more relaxed assumptions, compared with the Manhattan constraints.
Specifically, our method represents scenes as a signed distance function field, a color field, and a plane probability field, and optimizes these fields by volume rendering \cite{yariv2021volume} to reconstruct the scenes.
We obtain sparse depth by using the geometric constraints.
The geometric constraints project 3D points on the surface to similar-looking regions with similar features in different views, which ensures that corresponding points obtained by image feature matching represent the same 3D point on the surface.
Besides, we utilize plane constraints to make large planes in the scene parallel or vertical to the wall or floor.
We estimate large planes in images and impose the plane constraints by making the normal of planes parallel or orthogonal to the normal of the ground.
The plane constraints ensure smooth reconstruction for low-textured regions.

Our method is evaluated on the ScanNet \cite{dai2017scannet}.
Our method with sparse depth achieves comparable results with Manhattan-SDF with dense depth.
Our method with dense depth outperforms the Manhattan-SDF with dense depth.
Our method with dense depth achieves comparable results with existing methods that use 3D supervision.

In summary, our contributions are as follows:

$\bullet$ We propose a novel neural reconstruction method that uses sparse depth to achieve high quality reconstruction of indoor scenes without 3D supervision.

$\bullet$ We introduce the plane constraints to improve the reconstruction quality of planes that do not fit the Manhattan-world assumption. Our plane constraints are able to apply to more general planes.





\begin{figure*}
  \centering
  \subfloat{\includegraphics[width=0.90\textwidth]{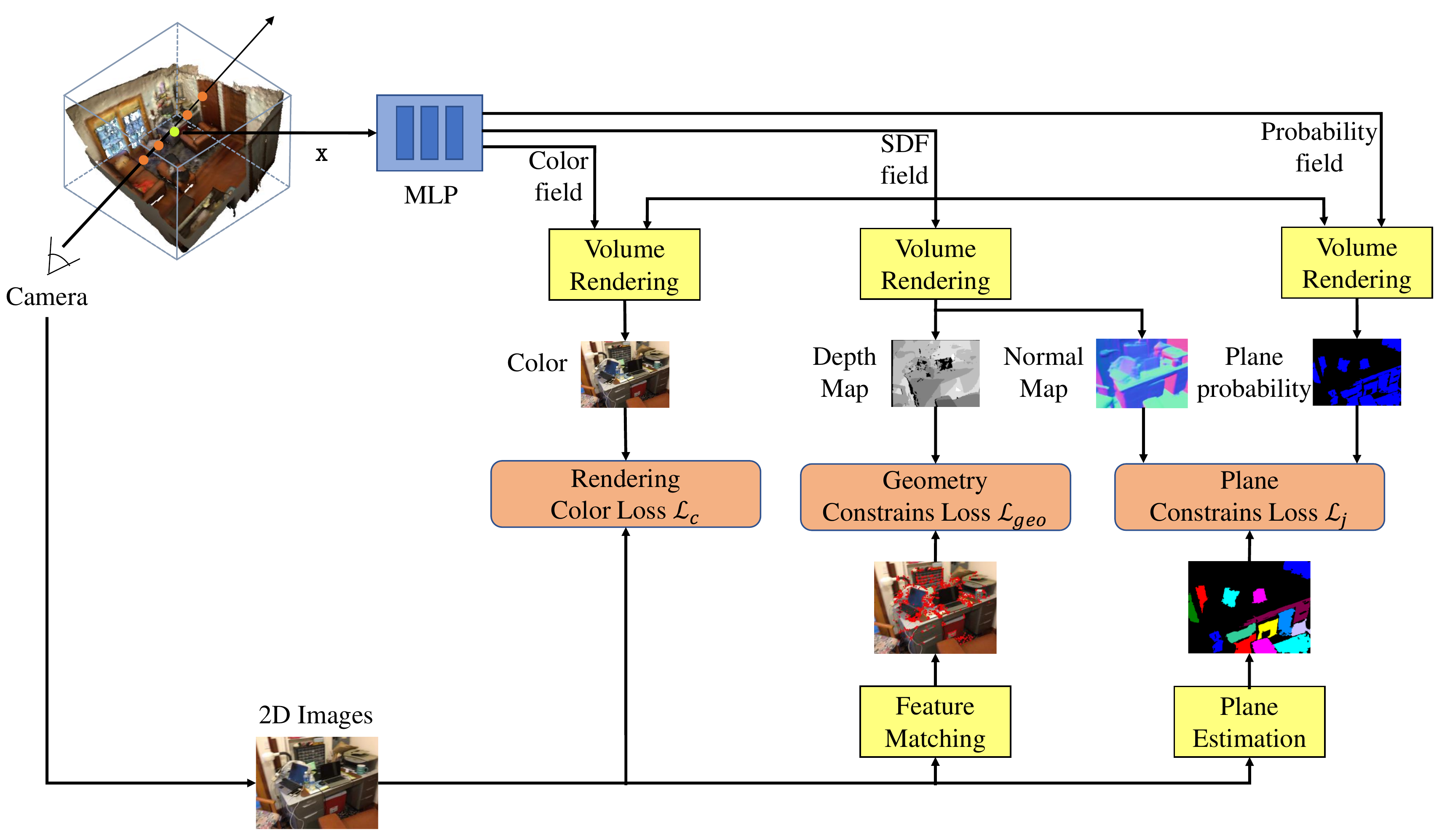}}
  \caption{\textbf{The overview of our method.} 
  }
  \label{figmethod}
 \end{figure*}

\section{Related Work}
Reconstructing 3D geometry structures of a scene from a sequence of images has been a longstanding computer vision problem. Traditional multi-view 3D scene reconstruction methods \cite{schonberger2016structure, schonberger2016pixelwise, seitz2006comparison} usually focus on estimating depth from a sequence of images, then fusing the depth maps and reconstructing the surface by screened Poisson surface reconstruction \cite{kazhdan2013screened}. Some works use deep neural networks to learn extracting features \cite{wang2021patchmatchnet, zagoruyko2015learning, suarez2020beblid}, matching features \cite{leroy2018shape, luo2016efficient, ummenhofer2017demon}, estimating depth maps \cite{huang2018deepmvs, yao2018mvsnet, yao2019recurrent, yu2020fast} or fusing depth maps \cite{donne2019learning, riegler2017octnetfusion} from a sequence of images. These deep methods improve performance compared with traditional 3D scene reconstruction methods. 
However, the fused reconstruction results are prone to be either layered or scattered, due to estimating depth of each key frame individually and estimation errors. 
To address the problem, neural scene reconstruction methods have been proposed. For example,
Atlas \cite{murez2020atlas} extracts the image features and regresses the TSDF of the scene. NeuralRecon \cite{sun2021neuralrecon} regresses TSDF directly using a coarse-to-fine framework which achieves real-time indoor scene reconstruction.
The TSDF fails to perform high-resolution reconstruction very well, because the TSDF representation is based on volume voxel.
To this end, some works use the neural implicit function to represent the scene, such as occupancy \cite{mescheder2019occupancy, peng2020convolutional, choy20163d, tatarchenko2017octree} and SDF \cite{park2019deepsdf, saito2019pifu}. 
The neural implicit representation is naturally continuous, which means free of the limitation of finite resolution.
These learning-based methods require 3D supervision for training, but obtaining it is often time-consuming and costly in real scenes. 
In contrast, our method reconstructs large-scale scenes without 3D supervision.

Several recent methods \cite{niemeyer2020differentiable, mildenhall2020nerf, wei2021nerfingmvs, oechsle2021unisurf} have demonstrated differentiable rendering for successfully reconstructing 3D scenes directly from 2D images. These methods can be divided into two groups based on the rendering technique: surface rendering and volume rendering. 
Surface rendering based methods, such as DVR \cite{niemeyer2020differentiable} and IDR \cite{yariv2020multiview}, achieve good performance in most cases, but these methods require pixel-accurate object masks for all images as input.
Therefore, these methods achieve unsatisfied performance for complex objects and scenes.
Volume rendering based methods, such as NeRF \cite{mildenhall2020nerf} and its variants \cite{wei2021nerfingmvs, deng2022depth, zhang2020nerf++}, render an image by learning alpha-compositing of a radiance field along rays. These methods have shown good performance on novel view synthesis. They encode a scene as continuous radiance fields of color and volume density, and map the position and view direction to an image by using differentiable ray marching.
Using volume density to represent the scene fails to extract high-quality surfaces well since the volume density representation does not have sufficient constraints on geometry. 
UNISURF \cite{oechsle2021unisurf}, NeuS \cite{wang2021neus} and VolSDF \cite{yariv2021volume} combine the surface representation such as occupancy values and signed distance function with the volume rendering to reconstruct the scene well and achieved good results. However, these methods may fail to handle complex geometry and large low-textured regions in large scenes. Differently, our method introduces geometry constraints and plane constraints to handle these regions in the large-scale indoor scene without 3D supervision.

Recently, some works focus on large-scale indoor scene reconstruction by using volume rendering, and achieve good results.
Manhattan-SDF \cite{guo2022neural} uses dense depth maps predicted by COLMAP \cite{schonberger2016structure} as supervision and uses Manhattan assumption to handle the walls and floor regions. 
However, obtaining dense depth maps is time-consuming by traditional methods without 3D supervision and the depth maps are noisy.
The plane reconstruction is poor in regions where the Manhattan World assumption does not hold, due to the noise in depth estimation.
To address these problems, NeuRIS \cite{wang2022neuris} utilizes dense normal maps predicted by a monocular method \cite{do2020surface} as supervision. 
MonoSDF \cite{yu2022monosdf} utilizes both depth and normal maps as supervision, and uses multi-resolution feature grids to help reconstruct the indoor scene. 
However, these two methods heavily rely on pre-trained networks supervised by 3D data.
In contrast to these methods, 
we perform differentiable volume rendering for scene reconstruction by using sparse depth under plane constraints without 3D supervision, to improve the plane reconstruction quality of scenes.

\section{Method}
\label{sec:method}

Given multi-view 2D images with camera poses, our goal is to reconstruct high-quality scenes without 3D supervised data. 
We represent a scene as a signed distance function (SDF) field, a color field, and a plane probability field, and optimize them by volume rendering.
We use geometry constraints to obtain sparse depth for improving the reconstruction quality of the regions that have complex geometry structures.
We estimate the plane of the scene and use plane constraints to improve the reconstruction quality of the large low-textured regions.
The overview of our method is illustrated in Figure \ref{figmethod}.

\subsection{Implicit Scene Representation}
\label{sec:Implicit Scene Representation}
We utilize two fields of SDF and color to model the scene geometry and scene appearance, respectively. We represent the scene geometry as an SDF field. 
The SDF takes a 3D point $\mathbf{x}\in\mathbb{R}^3$ as the input, and generates its distance $s(\mathbf{x})$ to the closest surface by using a  multi-layer perceptron (MLP) neural network  $f_{g}$, given by 
\begin{equation}
\begin{aligned}
\big(s(\mathbf{x}),\mathbf{z}(\mathbf{x})\big) = f_{g}(\mathbf{x};\theta_g),
\end{aligned}
\end{equation}
where $f_{g}$ is implemented as the MLP network with parameters $\theta_g$, and $\mathbf{z}(\mathbf{x})$ is the geometry feature calculated in \cite{yariv2020multiview}. The surface is defined as the zero level set of the SDF, that is,
\begin{equation}
\begin{aligned}
\quad{S} = \{\mathbf{x}|s(\mathbf{x})=0\}.\
\end{aligned}
\end{equation}
We represent the appearance of the scene as a color field. 
We define a network $f_{c}$ to predict the  RGB color $\mathbf{c}(\mathbf{x})$ for a 3D point $\mathbf{x}$ and a viewing direction $\mathbf{v}$, given by
\begin{equation}
\begin{aligned}
\mathbf{c}(\mathbf{x}) =f_{c}(\mathbf{x},\mathbf{v},\mathbf{n}(\mathbf{x}),\mathbf{z}(\mathbf{x});\theta_c),
\end{aligned}
\end{equation}
where $\mathbf{n}(\mathbf{x})$ is the unit normal obtained by computing the gradient of our SDF function $f_{g}$, $\mathbf{z}(\mathbf{x})$ is the geometry feature of the output of the MLP, and $\theta_c$ is the parameters of the network.

\subsection{Volume Rendering of Implicit Surfaces}
\label{sec:Volume Rendering of Implicit Surfaces}
Following \cite{yariv2021volume, wang2021neus}, we apply differentiable volume rendering to optimize the scene representation from images. Specifically, we cast a ray $\mathbf{r}=\mathbf{o}+t\mathbf{v}$ with the origin in the camera center $\mathbf{o}$ and the viewing direction $\mathbf{v}$ to render a pixel of the image. 
We sample $N$ points $\{\mathbf{x}_i|\mathbf{x}_i=\mathbf{o}+t_i\mathbf{v},i=1,2,3...,N\}$ along the camera ray and predict the $s(\mathbf{x}_i)$ and $\mathbf{c}(\mathbf{x}_i)$ for each point. For convenience, here we use the $s_i$ and $\mathbf{c}_i$ to represent the $s(\mathbf{x}_i)$ and $\mathbf{c}(\mathbf{x}_i)$.
We transform the SDF $s_i$ to the volume density $\sigma_i$ by

\begin{equation}
\sigma(s)=
\begin{cases}
\frac{1}{2\beta}\text{exp}\left(\frac{s}{\beta}\right)    & \text{if}\quad s \leq 0, \\
\frac{1}{\beta}\left(1- \frac{1}{2}\text{exp}\left(-\frac{s}{\beta}\right) \right) & \text{if}\quad s > 0,
\end{cases}
\end{equation}
where $\beta$ is a learnable parameter. 
We accumulate the color $\hat{C}(\mathbf{r})$ along the ray $\mathbf{r}$ via numerical quadrature \cite{mildenhall2020nerf}:
\begin{equation}
\begin{aligned}
&\hat{C}(\mathbf{r})=\sum_{i=1}^{N}T_{i}(1-\text{exp}(-\sigma_{i}\delta_{i}))\mathbf{c}_{i},\\
& \text{where} \quad T_{i}=\text{exp}\left(-\sum_{j=1}^{i-1}\sigma_{j}\delta_{j}\right),
\end{aligned}
\end{equation}
and $\delta_{i}$ is the distance between adjacent samples.

\subsection{Geometry Constraints}
\label{sec:geometry constraints}
\begin{figure}
  \centering
  \subfloat{\includegraphics[width=0.42\textwidth]{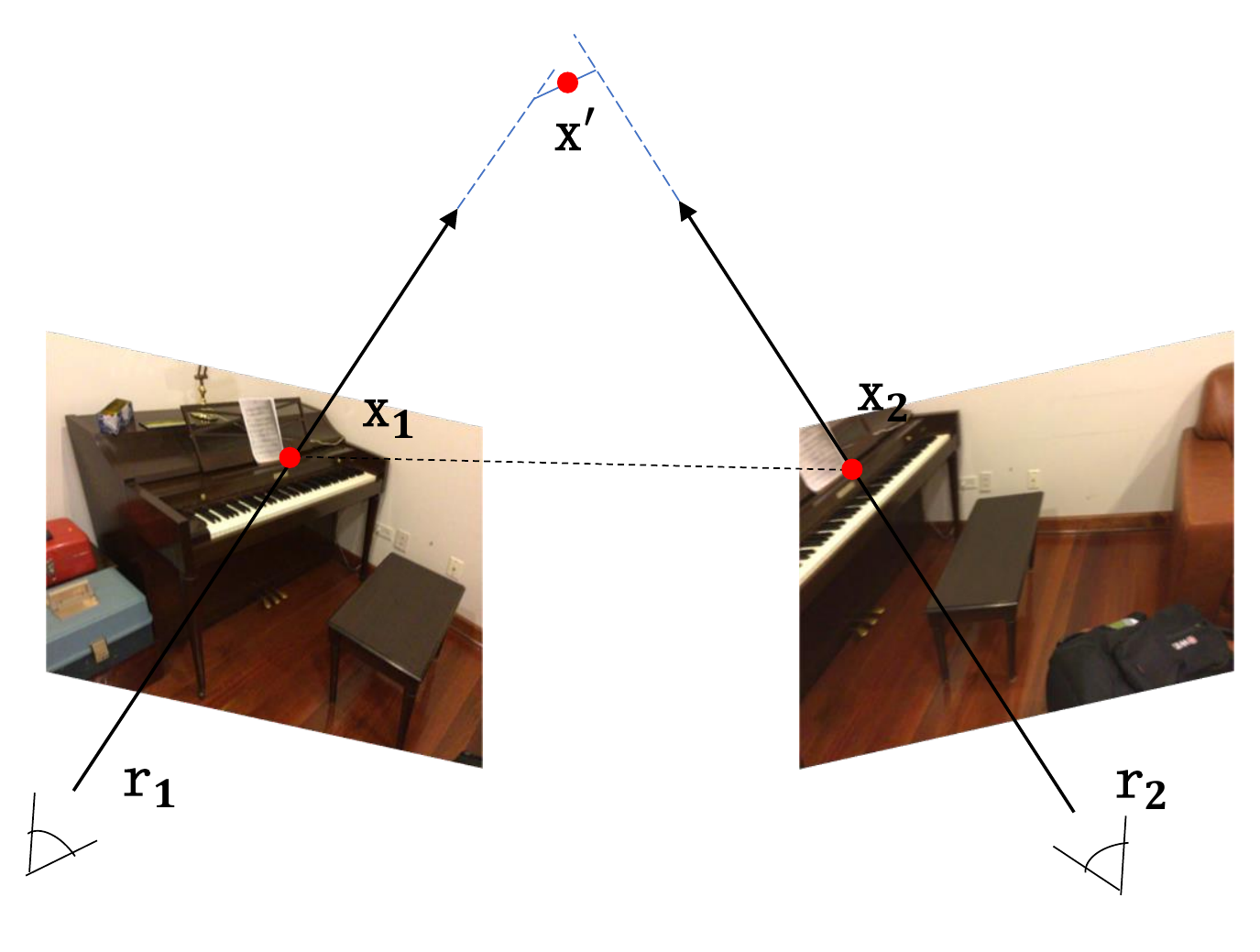}}
  \caption{\textbf{The illustration of geometry constraints.} Given two images, we match the feature points $\mathbf{x}_1$ and $\mathbf{x}_1$. We cast rays $\mathbf{r}_1$ and $\mathbf{r}_2$ from the camera center through the matched points. The midpoint $\mathbf{x}^{\prime}$ of the common perpendiculars of rays is on the scene surface.}
  \label{fig3}
 \end{figure}

We utilize geometry constraints to obtain sparse depth for helping reconstruct the regions with complex geometry structures of the indoor scene. 
We observe that (1) the 3D point on the surface is projected to similar-looking regions with similar features in different views, as shown in Figure \ref{fig3}; (2) the regions with complex geometry structures always have sharp features. These two observations inspire us to use geometric constraints to generate sparse  depth for improving the reconstruction quality of the regions with complex geometry.

As illustrated in Figure \ref{fig3}, given multi-view images, we first extract the feature points (e.g., ORB feature points \cite{rublee2011orb} or SIFT feature points \cite{lowe2004distinctive}) of the images, and then we obtain the point correspondences by matching the feature points in adjacent images. We cast a ray from the camera center through the pixel of each matching point. 
For one correspondence of two points $\mathbf{x}_1$ and $\mathbf{x}_2$, two rays $\mathbf{r}_1$ and $\mathbf{r}_2$ will intersect at a point on the surface of the scene, theoretically,
but they usually do not intersect due to the existence of errors. Therefore, we calculate the point closest to the two lines as an approximate intersection point.
For the two rays, we calculate their common perpendiculars, and the approximate intersection point is the midpoint $\mathbf{x}^{\prime}$ of the line between the common perpendiculars and the actual intersection point of the two rays.
The approximate depth $D_{app}$ of the two rays can be obtained by projecting the approximate intersection points on the two rays respectively. If the distance between the two rays is larger than a threshold, we consider that this correspondence is wrong and discard it.

For the scene representation, we calculate the depth $\hat{D}(\mathbf{r})$ from a viewpoint by using the volume rendering:
\begin{equation}
\begin{aligned}
&\hat{D}(\mathbf{r})=\sum_{i=1}^{N}T_{i}(1-\text{exp}(-\sigma_{i}\delta_{i}))t_{i}.\\
\end{aligned}
\end{equation}
We use a geometry loss function $\mathcal{L}_{geo}$ of matching points to assist the learning of the textured regions,
\begin{equation}
\begin{aligned}
\mathcal{L}_{geo}=\sum_{\mathbf{r}\in\mathcal{M}}|&\hat{D}(\mathbf{r}) - D_{app}(\mathbf{r})|,
\end{aligned}
\end{equation}
where $\mathcal{M}$ is the set of the camera rays along the matching points and $|\cdot|$ is the absolute value. This geometry loss improves the reconstruction quality of the regions which have sharp features.

\subsection{Plane Constraints}
\label{sec:plane constraints}
\begin{figure}
  \subfloat[]{\includegraphics[width=0.11\textwidth]{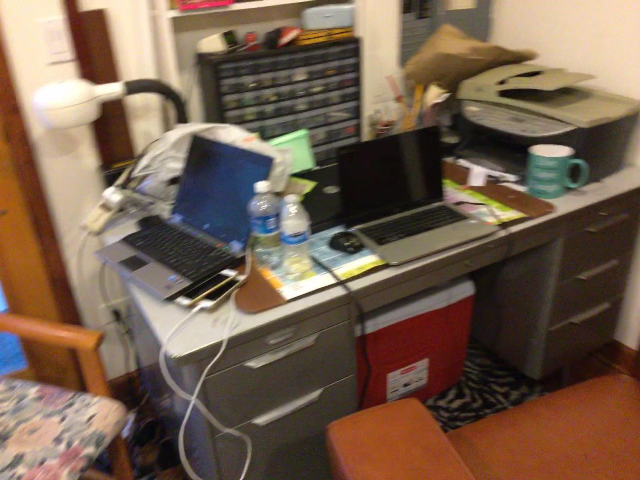}}
  \hspace{0.05mm}
  \subfloat[]{\includegraphics[width=0.11\textwidth]{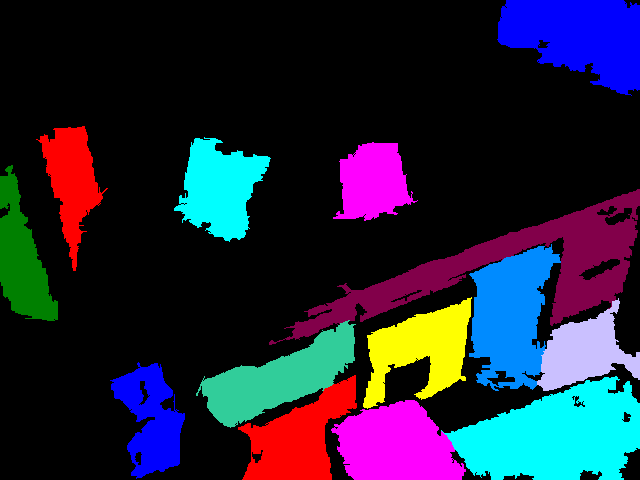}}
  \hspace{0.05mm}
  \subfloat[]{\includegraphics[width=0.11\textwidth]{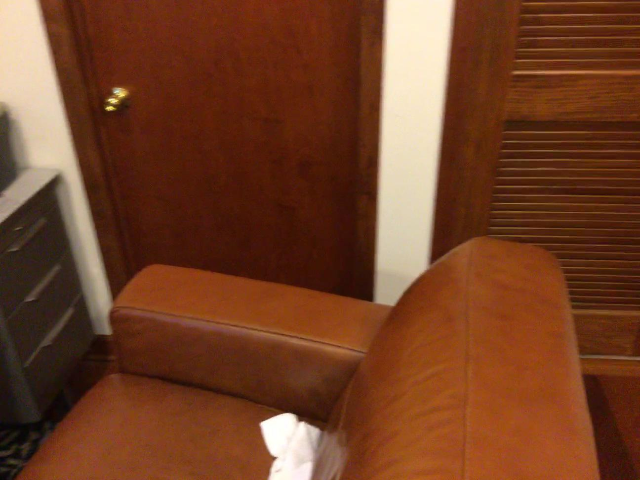}}
  \hspace{0.05mm}
  \subfloat[]{\includegraphics[width=0.11\textwidth]{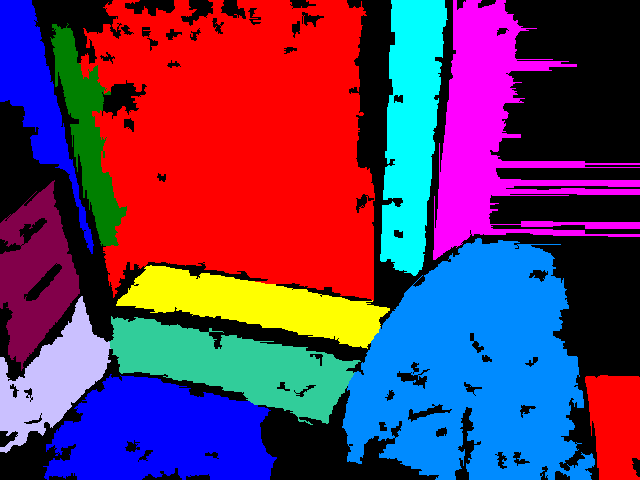}}
  \caption{\textbf{Two examples of plane estimation.} (a) and (c) are the input images, (b) and (d) are the results of plane estimation. 
  The colored regions represent the large planes in the scene that need to be imposed by plane constraints.
  The black regions do not contain large planes.
  We can see that not only walls and floors, but also the surfaces of tables have been successfully estimated.}
  \label{fig4}
 \end{figure}

We utilize the plane constraints to help reconstruct the low-textured plane regions.
We observe that the large low-textured plane regions are located not only on floors and walls, but also lie on tables and beds, as shown in Figure \ref{fig4}.
These regions are often parallel or vertical to the floor. For example, the tabletop and bed in Figure \ref{fig4} are both parallel to the floor.
The wall, the side of the table, the bookshelf, and other planes are all perpendicular to the floor.

According to the observation, we apply the plane constraints to help reconstruct the large low-textured regions. We utilize the Felzenswalb superpixel segmentation  algorithm \cite{felzenszwalb2004efficient} to obtain the plane regions. The algorithm follows a greedy approach and segments areas with low gradients, and produces more plane regions.
The plane segmentation results contain many small planes which are discarded in the plane constraints. 
We only keep the planes which are larger than a certain proportion of the image to impose the constraints.
We calculate the normal $\hat{N}(\mathbf{r})$ from a viewpoint by using the volume rendering:
\begin{equation}
\begin{aligned}
&\hat{N}(\mathbf{r})=\sum_{i=1}^{N}T_{i}(1-\text{exp}(-\sigma_{i}\delta_{i}))\mathbf{n}_{i},\\
&\text{where} \quad \mathbf{n}_{i} = \nabla {f_{g}(\mathbf{x}_i;\theta_g)},
\end{aligned}
\end{equation}
and $\nabla {f_{g}(\mathbf{x}_i;\theta_g)}$ is the spatial gradient of SDF. We assume the floors are vertical to the z-axis. We design a plane loss function $\mathcal{L}_{pla}$ that enforces the normal of large regions to be parallel or orthogonal to the upper unit vector as
\begin{equation}
\begin{aligned}
\mathcal{L}_{pla}=\sum_{\mathbf{r}\in\mathcal{P}}\min_{i\in\{-1,0,1\}}|i - \hat{N}(\mathbf{r})\cdot\mathbf{n}_f|,
\end{aligned}
\end{equation}
where $\mathcal{P}$ is the set of camera rays of images pixels that are segmented as the large plane regions, and $\mathbf{n}_f=(0,0,1)$ is the upper unit vector that denotes the normal of floors.

We use an MLP network $f_{p}$ to denote the probability of a point on the large plane. The probability logits are defined as
\begin{equation}
\begin{aligned}
p(\mathbf{x}) = f_{p}(\mathbf{x};\theta_p),
\end{aligned}
\end{equation}
where $\theta_p$ is the parameters of the MLP. For each image, we render the probability logits similar to image rendering as
\begin{equation}
\begin{aligned}
&\hat{P}(\mathbf{r})=\sum_{i=1}^{N}T_{i}(1-\text{exp}(-\sigma_{i}\delta_{i})){p}_{i}.
\end{aligned}
\end{equation}
We use a joint optimization loss to jointly optimize the scene representation and plane region estimation results, that is
\begin{equation}
\begin{aligned}
&\mathcal{L}_{j}=\sum_{\mathbf{r}\in\mathcal{P}}\hat{P}(\mathbf{r})\mathcal{L}_{pla}(\mathbf{r})+\mathcal{L}_{p},\\
&\text{where} \quad \mathcal{L}_{p}=-\sum_{\mathbf{r}\in\mathcal{R}}P(\mathbf{r})\log{\hat{P}(\mathbf{r})},
\end{aligned}
\end{equation}
and $\hat{P}(\mathbf{r})$ is the rendered probability, $P(\mathbf{r})$ is the plane obtained by the Felzenswalb. $\quad \mathcal{L}_{p}$ is the cross entropy loss to avoid the $\hat{P}(\mathbf{r})$ converging to zero.

\subsection{Training}
During the training stage, we sample a batch of pixels and minimize the loss functions of the color and the constraints. 
The overall loss is defined as
\begin{equation}
\begin{aligned}
\mathcal{L} = \lambda_c\mathcal{L}_c+\lambda_{geo}\mathcal{L}_{geo}+\lambda_{j}\mathcal{L}_{j}+\lambda_{eik}\mathcal{L}_{eik}.
\end{aligned}
\end{equation}
The color loss $\mathcal{L}_c$ is defined as
\begin{equation}
\begin{aligned}
\mathcal{L}_c=\sum_{\mathbf{r}\in\mathcal{R}}\big|\big|&\hat{C}(\mathbf{r})-C(\mathbf{r})\big|\big|,
\end{aligned}
\end{equation}
where $\mathcal{R}$ is the set of sample pixel, $C(\mathbf{r})$ is the ground truth pixel color, and $||\cdot||$ is the 1-norm.

The Eikonal loss \cite{gropp2020implicit} is introduced to regularize SDF values in 3D space:
\begin{equation}
\begin{aligned}
\mathcal{L}_{eik}=\sum_{\mathbf{x}\in\mathcal{X}}(||\nabla {f_{g}(\mathbf{x};\theta_g)}||_2-1)^2,
\end{aligned}
\end{equation}
where $\mathcal{X}$ are a set of uniform sampling points and near surface points, and $||\cdot||_2$ is the 2-norm.

\section{Implementation Details}
We implement our method in PyTorch \cite{paszke2019pytorch} and use Adam optimizer \cite{kingma2014adam} with a learning rate of $5\rm{e}$-4. We sample 1024 rays from pixel points for each batch to train the network. The network is trained for 50k iterations on one NVIDIA RTX3090 GPU. We use sphere initialization \cite{atzmon2020sal} to initialize the network parameters. 
In the early stage of training, we aim to reconstruct the structure of the scene first, so we tend to select matching pixel points for training and set a larger geometry weight $\lambda_{geo}$. As the training process goes on, we reduce the weight of the geometric loss and random sample pixel points for training. We use Marching Cubes algorithm \cite{lorensen1987marching} to extract surface mesh from the learned signed distance function.

\section{Experiments}

\begin{figure*}
\centering
  \captionsetup[subfloat]{labelsep=none,format=plain,labelformat=empty}
  \subfloat[VolSDF]{\includegraphics[width=0.19\textwidth]{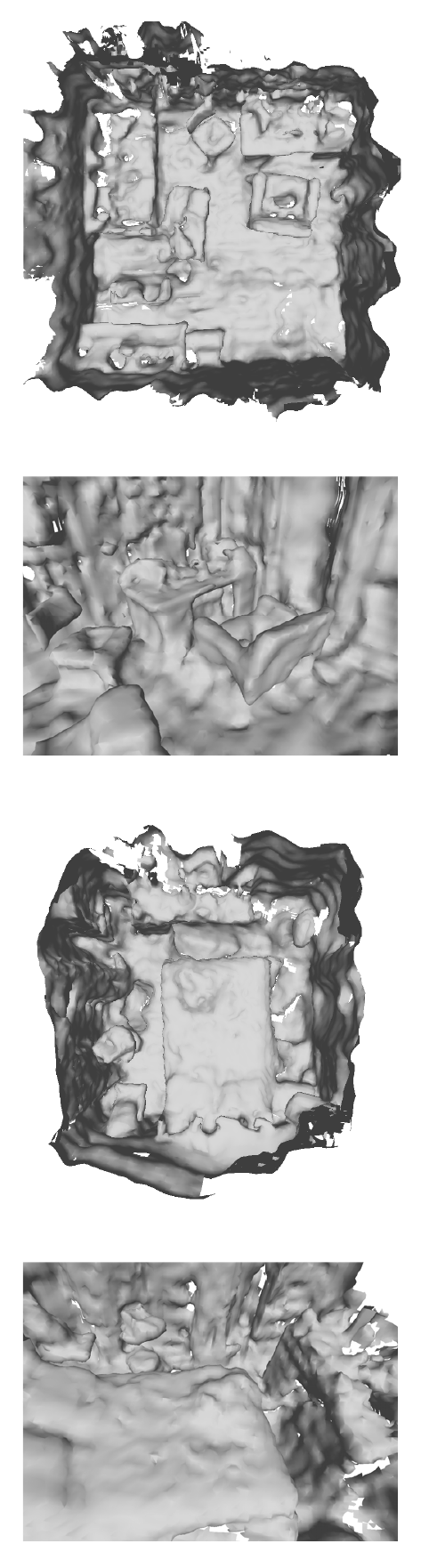}}
  \subfloat[VolSDF-G]{\includegraphics[width=0.19\textwidth]{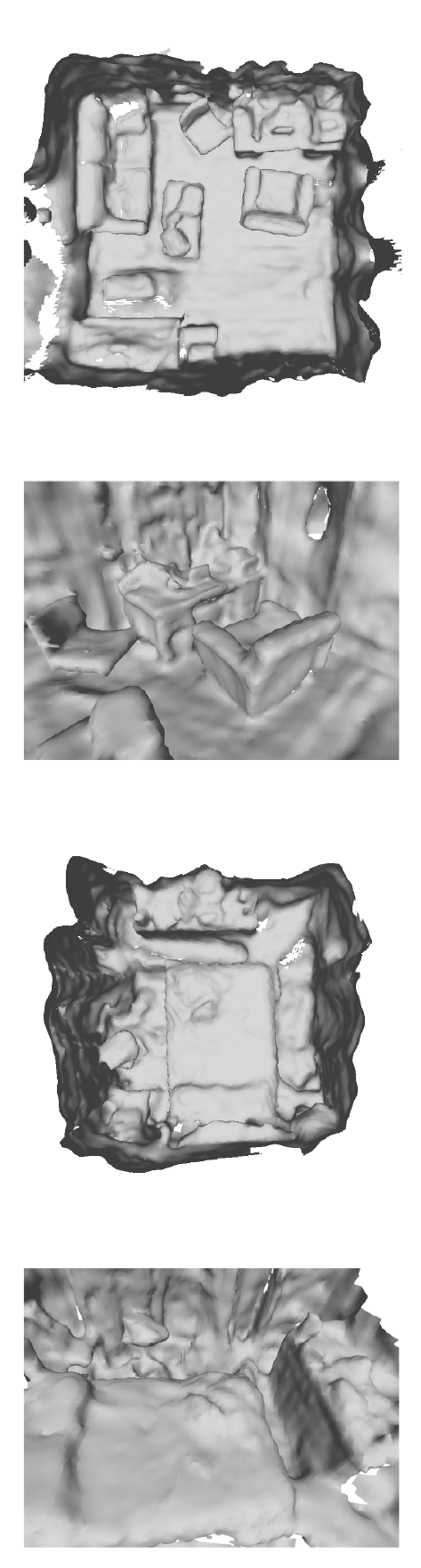}}
  \subfloat[VolSDF-P]{\includegraphics[width=0.19\textwidth]{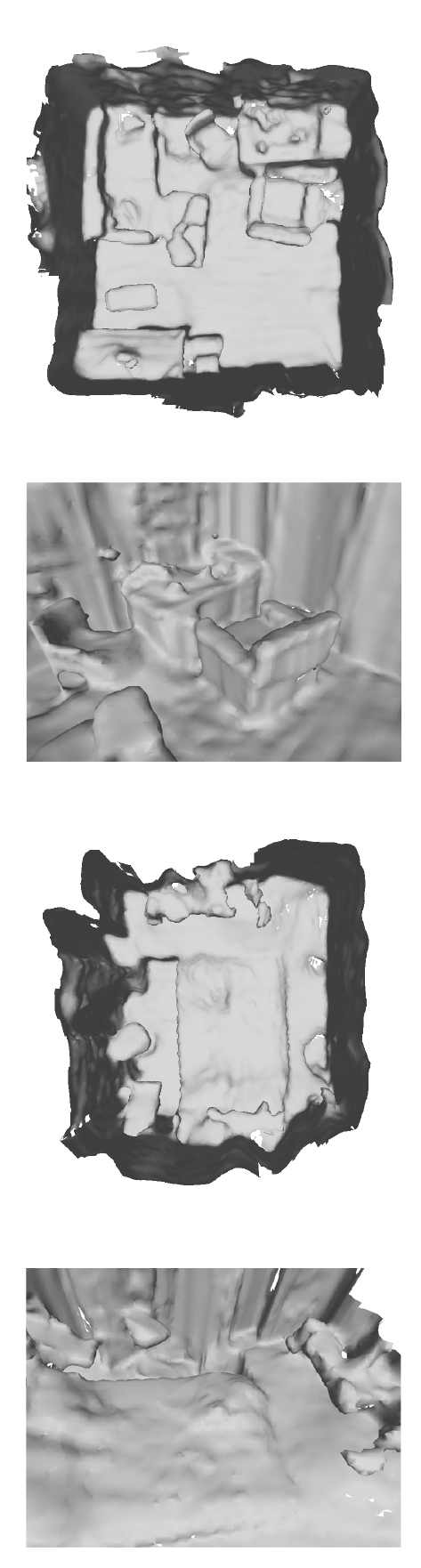}}
  \subfloat[Ours]{\includegraphics[width=0.19\textwidth]{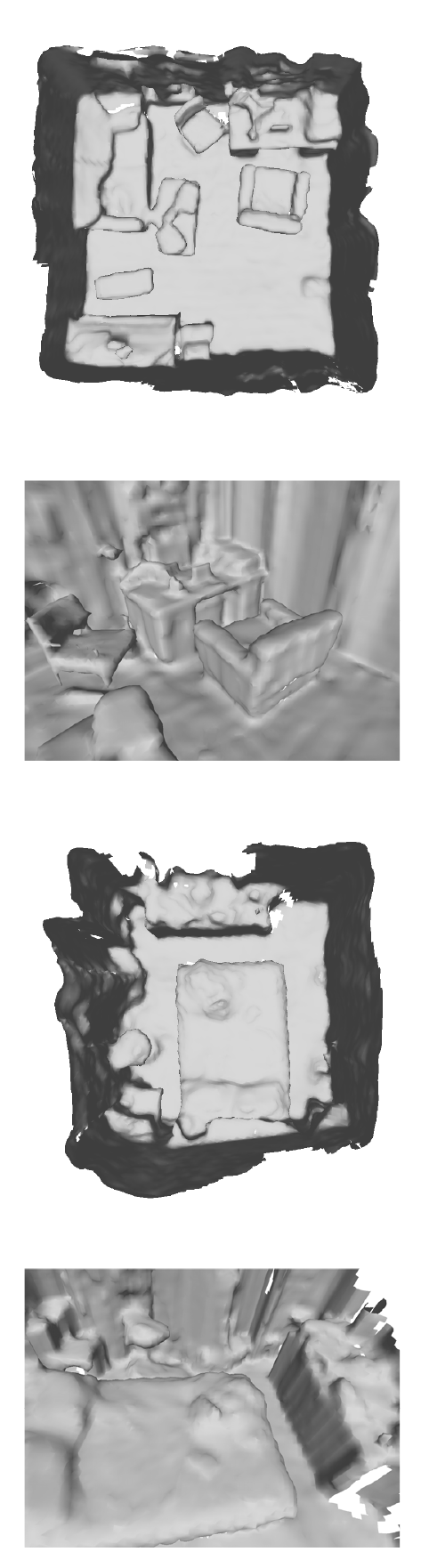}}
  \subfloat[Ground Truth]{\includegraphics[width=0.19\textwidth]{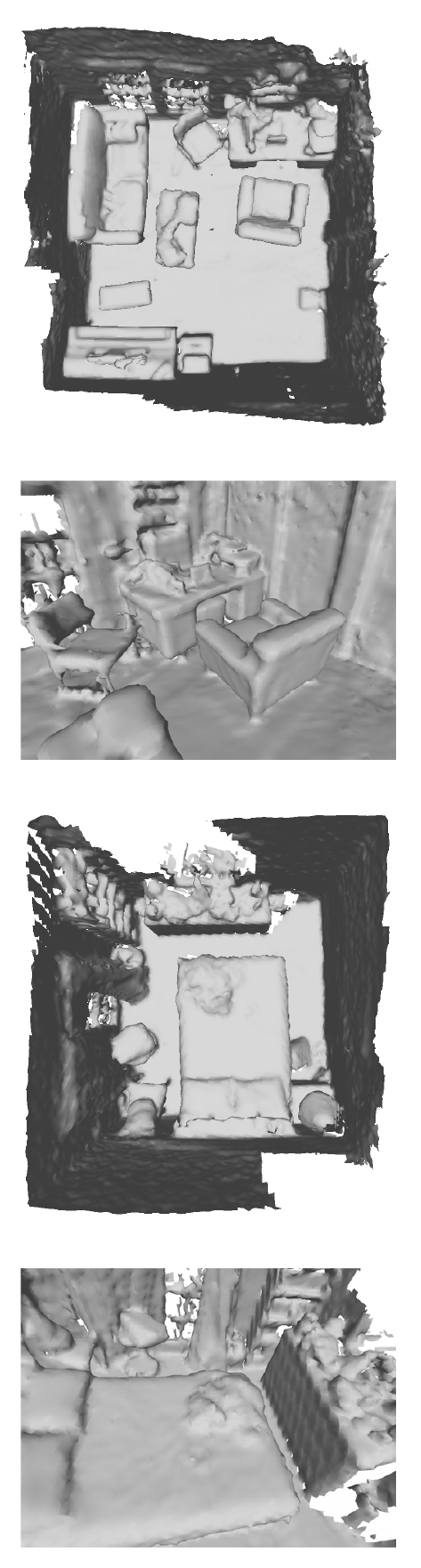}}
  \caption{\textbf{Qualitative results of ablation studies on ScanNet.} Our method produces much more accurate and smooth reconstruction results compared with our baselines, i.e., VolSDF, VolSDF-G, and VolSDF-P. VolSDF use 2D images to reconstruct the scenes without constraints.VolSDF-G uses the geometry constraints to reconstruct more accurate results than the VolSDF. VolSDF-P uses the plane constraints to reconstruct smoother and more complete planes compared with VolSDF. Our method reconstructs more accurate and smoother results than VolSDF-G and VolSDF-P, by imposing the two constraints.}
  \label{Ablation}
 \end{figure*}

\noindent \textbf{Dataset.} We perform the experiments on the ScanNet(V2) \cite{dai2017scannet}, which contains 1613 indoor scenes.
It provides ground truth camera poses of 2D images, surface reconstructions and instance-level semantic segmentations of the scenes. For each scene, it contains 1K-5K RGB-D images, and we uniform sample one tenth images for reconstruction. The experiment setup is the same as the work of Guo \emph{et al.}  \cite{guo2022neural}. 

\noindent \textbf{Metrics.} We evaluate the 3D surface geometry results using 5 standard metrics defined in \cite{murez2020atlas}: accuracy, completeness, precision, recall, and F-score. Among these metrics, F-score is usually considered as the most suitable metric to measure 3D reconstruction quality according to the work of Sun \emph{et al.} \cite{sun2021neuralrecon}.
The definitions of these metrics are detailed in the \textit{supplementary material}.

\begin{table}
   \small
   \centering
   \begin{tabular}{c|c|c|c|c|c}
    \hline
    \bfseries Method & Acc$\downarrow$ & Comp$\downarrow$ & Pre$\uparrow$ & Recall$\uparrow$ & \bfseries F-score$\uparrow$  \\
    \hline
    VolSDF & $0.414$ & $0.120$ & $0.321$ &$0.394$ & $0.346$ \\
    VolSDF-G & $0.106$ & $0.101$ & $0.507$ &$0.469$ & $0.487$\\
    VolSDF-P & $0.100$ & $0.118$ & $0.493$ &$0.427$ & $0.459$\\
    \hline
    Ours & $\mathbf{0.068}$ & $\mathbf{0.079}$ & $\mathbf{0.626}$ &$\mathbf{0.551}$ & $\mathbf{0.586}$\\
    
    \hline
   \end{tabular}
   \caption{\textbf{Quantitative results of ablation studies on ScanNet.} Our method significantly improves the accuracy and completeness compared of our baselines of VolSDF, VolSDF-G and VolSDF-P.}
   \label{table:ablation studies}
  \end{table}

\begin{table*}
   \centering
   \setlength{\tabcolsep}{5mm}
   \resizebox{15cm}{!}{\begin{tabular}{c|c|c|c|c|c|c|c}
    \hline
    \bfseries 3D & \bfseries Method & \bfseries dense & Acc$\downarrow$ & Comp$\downarrow$ & Pre$\uparrow$ & Recall$\uparrow$ & \bfseries F-score$\uparrow$  \\
    \hline
    \multirow{10}{*}{\XSolidBrush}   
    & NeRF \cite{mildenhall2020nerf}  & \XSolidBrush & $0.735 $ & $0.177 $ & $0.131 $ & $0.290 $ & $0.176 $\\
    & UNISURF \cite{oechsle2021unisurf}  & \XSolidBrush & $0.554 $ & $0.164 $ & $0.212 $ & $0.362 $ & $0.267 $\\
    & NeuS \cite{wang2021neus}  & \XSolidBrush & $0.179 $ & $0.208 $ & $0.313 $ & $0.275 $ & $0.291 $\\
    & VolSDF \cite{yariv2021volume} & \XSolidBrush & $0.414 $ & $0.120 $ & $0.321 $ & $0.394 $ & $0.346 $\\    
    & Manhattan-SDF-s \cite{guo2022neural} & \XSolidBrush & $0.076 $ & $0.079 $ & $0.577 $ & $0.506 $ & $0.541 $\\
    & \textbf{Ours} & \XSolidBrush & $\mathbf{0.068}$ & $\mathbf{0.079}$ & $\mathbf{0.626}$ &$\mathbf{0.551}$ & $\mathbf{0.586}$\\
    \cline{2-8}
    & COLMAP \cite{schonberger2016structure} & \Checkmark & $\mathbf{0.047} $ & $0.235 $ & $0.711 $ & $0.441 $ & $0.537 $ \\  
    & Manhattan-SDF \cite{guo2022neural} & \Checkmark & $0.072 $ & $0.068 $ & $0.621 $ & $0.586 $ & $0.602 $\\ 
    & \textbf{Ours-d} & \Checkmark & $0.057$ & $\mathbf{0.057}$ & $\mathbf{0.708}$ &$\mathbf{0.677}$ & $\mathbf{0.692} $\\
    \hline    
    \multirow{2}{*}{\Checkmark} &Atlas \cite{murez2020atlas} & \Checkmark & $0.124 $ & $0.074 $ & $0.413 $ & $\mathbf{0.711} $ & $0.520 $\\
    & NeuRIS \cite{wang2022neuris} & \Checkmark & $\mathbf{0.050} $ & $\mathbf{0.049} $ & $\mathbf{0.717} $ & $0.669 $ & $\mathbf{0.692} $\\
    
    \hline
   \end{tabular}}
   \caption{\textbf{Quantitative results of our method and state-of-the-art mthods on ScanNet.} The first column means whether the method uses 3D data or pre-trained networks supervised by 3D data. 
   The third column means whether the method uses dense depth or normal maps.
   Our method with sparse depth outperforms the existing method using sparse depth.}
   \label{table:comparisons}
  \end{table*}

\noindent \textbf{Competing methods.} The competing methods include: (1) Classical MVS method: COLMAP \cite{schonberger2016structure}, we use ground truth camera poses to reconstruct the point cloud, and use Screened Poisson Surface Reconstruction \cite{kazhdan2013screened} to reconstruct mesh from a point cloud. (2) TSDF based method: Atlas \cite{murez2020atlas}. Atlas directly regresses a TSDF from a set of posed RGB images. (3) Neural volume rendering based methods: NeRF \cite{mildenhall2020nerf}, UNISURF \cite{oechsle2021unisurf}, NeuS \cite{wang2021neus} and VolSDF \cite{yariv2021volume}. These methods do not use 3D data. (4) State-of-the-art neural scene reconstruction methods: Manhattan-SDF \cite{guo2022neural}, NeuRIS \cite{wang2022neuris}. 
For neural volume rendering based
methods and neural scene reconstruction methods, we extract mesh by using Marching Cubes algorithm\cite{lorensen1987marching}.

\subsection{Ablation Studies}  

We conduct ablation studies on ScanNet to verify the effectiveness of each component in our method.
We conduct experiments in four different settings: (1) Baseline setting of VolSDF \cite{yariv2021volume}, we train networks only with 2D image supervision. 
(2) VolSDF-G, we add the sparse depth obtained by geometry constraints to the VolSDF. (3) VolSDF-P, we add the plane constraints to the VolSDF. (4) Ours, we learn the scene with both sparse depth and the plane constraints.
The quantitative results are shown in Table \ref{table:ablation studies}, and the qualitative results are shown in Figure \ref{Ablation}.

The comparison between VolSDF and VolSDF-G shows that the sparse depth obtained by geometry constraints in our method improves $0.141$ precision in terms of F-score, as shown in Table \ref{table:ablation studies}.
The sparse depth obtained by geometry constraints helps reconstruct the regions that have complex geometry structures, but the reconstruction quality is still poor  for the large low-textured regions, as shown in VolSDF and VolSDF-G in Figure \ref{Ablation}.

The comparison between VolSDF and VolSDF-P shows that the plane constraints in our method improve $0.113$ precision in terms of F-score, as shown in Table \ref{table:ablation studies}.
The plane constraints help reconstruct the large low-textured regions, such as the wall, floor and table. Nevertheless, the reconstruction quality is still poor  for the regions with sharp features, as shown in VolSDF and VolSDF-P in Figure \ref{Ablation}.

Our method improves $0.240$ precision in terms of F-score compared with VolSDF. 
Our method combines the sparse depth and plane constraints, and reconstructs both sharp features regions and large low-textured regions well. 

\subsection{Comparisons}

\begin{figure*}
\centering
  \subfloat{\rotatebox{90}{\scriptsize{~~~~~~~COLMAP \cite{schonberger2016structure}}}
  \includegraphics[width=0.95 \textwidth]{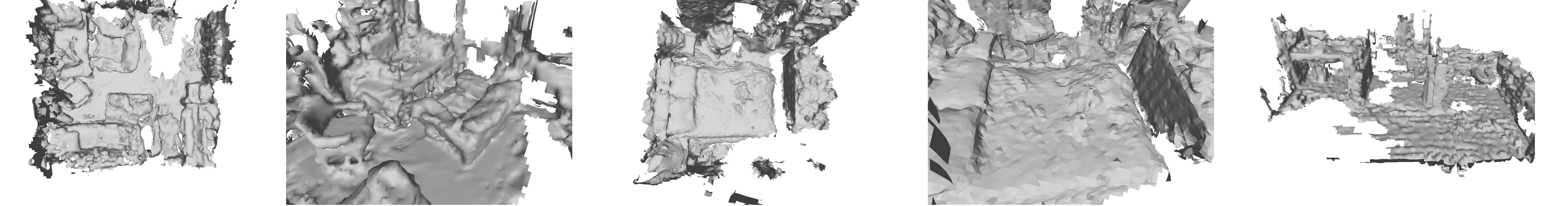}}\\
  \subfloat{
  \rotatebox{90}{\scriptsize{~~~~~~~~NeuRIS \cite{wang2022neuris}}}\includegraphics[width=0.95\textwidth]{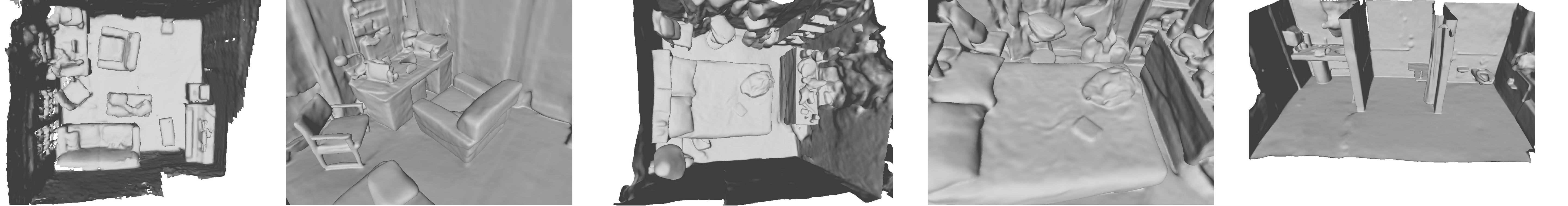}}\\
  \subfloat{
  \rotatebox{90}{\scriptsize{~~~~Manhattan-SDF-s }}\includegraphics[width=0.95\textwidth]{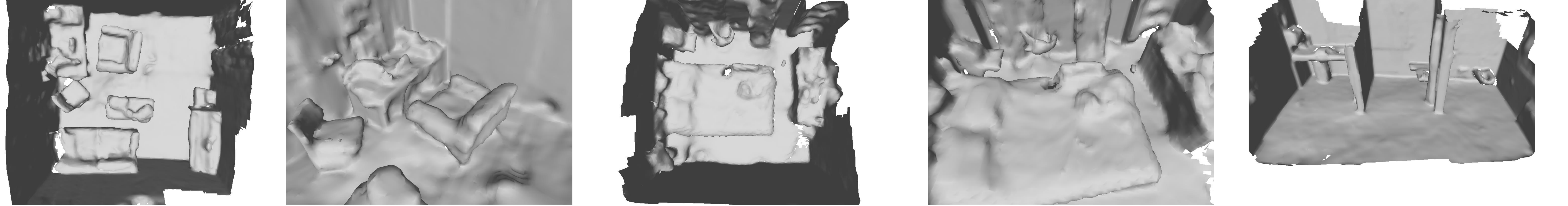}}\\
  \subfloat{
  \rotatebox{90}{\scriptsize{Manhattan-SDF \cite{guo2022neural}}}\includegraphics[width=0.95\textwidth]{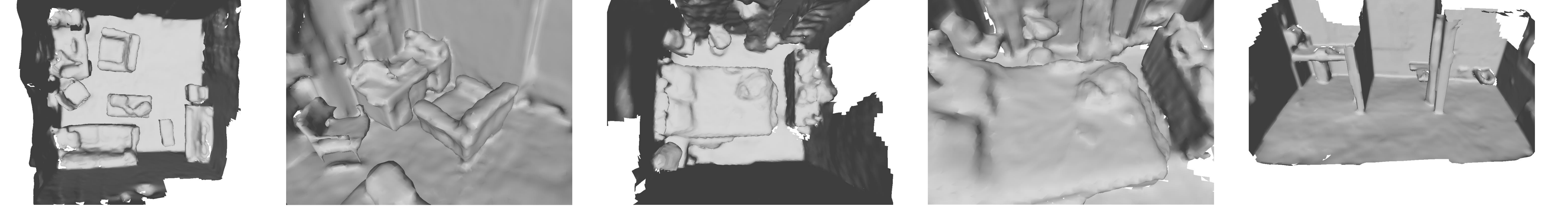}}\\
  \subfloat{\rotatebox{90}{\scriptsize{~~~~~~~~~~~~~~Ours}}
  \includegraphics[width=0.95\textwidth]{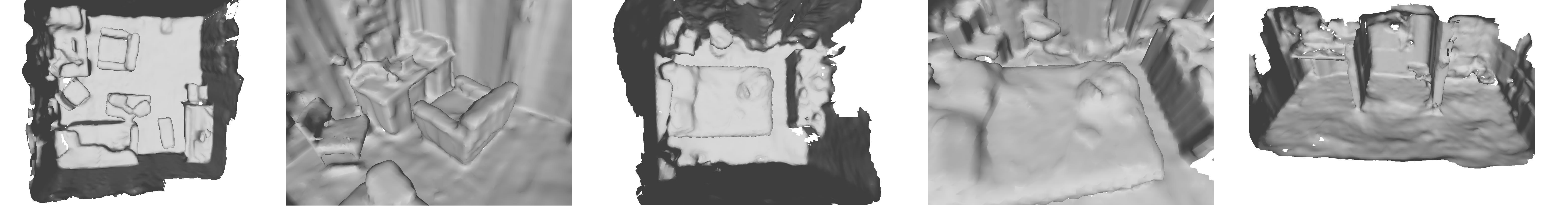}}\\
  \subfloat{\rotatebox{90}{\scriptsize{~~~~~~~~~~~~~~Ours-d}}
  \includegraphics[width=0.95\textwidth]{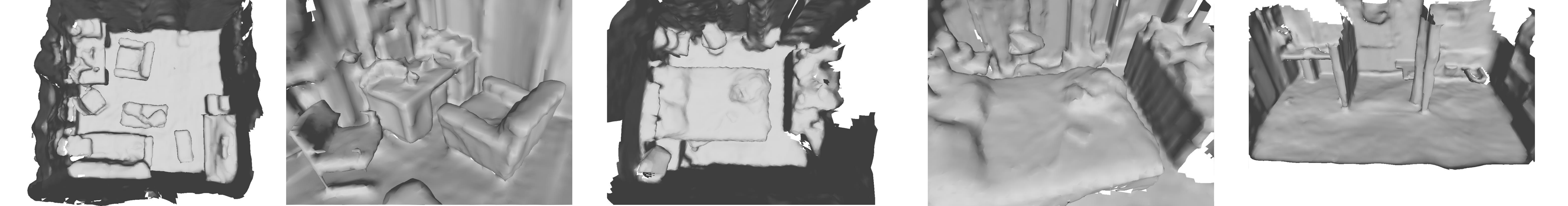}}\\
  \subfloat{
  \rotatebox{90}{\scriptsize{~~~~~~Ground Truth}}\includegraphics[width=0.95\textwidth]{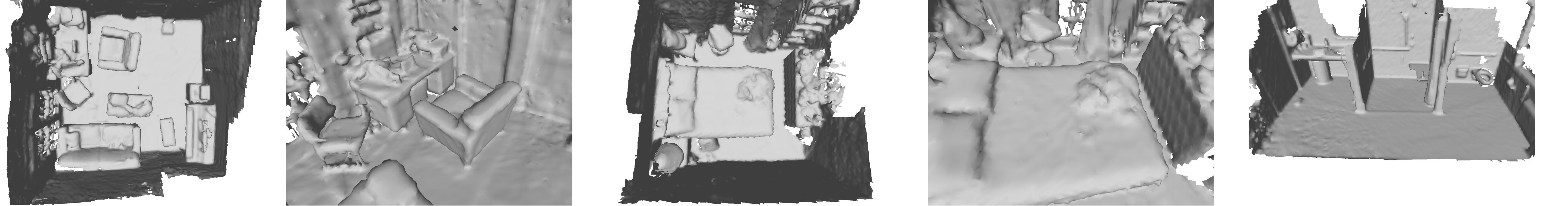}}
  \caption{\textbf{Qualitative results of our method and sate-of-the-art methods on ScanNet.} Our method outperforms COLMAP and Manhattan-SDF methods, and performs comparable results with NeuRIS with pre-trained networks. Our method produces accurate results in the regions of complex structures and coherent results in the regions of large planes.}
  \label{fig:Comparisons}
 \end{figure*}

We compare our method with state-of-the-art reconstruction methods on ScanNet. The averaged quantitative results of the geometry quality are shown in Table \ref{table:comparisons}. The qualitative results are shown in the Figure \ref{fig:Comparisons}. 
The individual scene results are in the \textit{supplementary material}.

We categorize existing methods into two types  according to whether they use 3D supervision, as shown in Table \ref{table:comparisons}.
We first compare our method with existing methods that do not use 3D supervision and do not need dense depth. The performance of NeRF \cite{mildenhall2020nerf} is poor, 
because its volume density representation of the scene has geometry ambiguity. UNISURF \cite{oechsle2021unisurf}, Neus \cite{wang2021neus} and VolSDF \cite{yariv2021volume} represent the scene as occupancy and SDF to  improve the reconstruction of  the scene surface. However, they do not perform well in indoor scenes. 
Manhattan-SDF \cite{guo2022neural} with sparse depth, \textit{i.e.}, Manhattan-SDF-s, has achieved performance improvement, but compared with Manhattan-SDF \cite{guo2022neural}, the reconstruction quality of Manhattan-SDF-s is significantly reduced because of the lack of sufficient constraints on the planes. 
Our method with sparse depth outperforms these methods in indoor scenes, which proves the superiority of our method.
Then, we compare our method with existing methods using dense depth without 3D supervision.
COLMAP \cite{schonberger2016structure} achieves high accuracy and precision, because it filters out inconsistent reconstruction points between multiple views in the fusion phase. Its completeness and recall are low.
Manhattan-SDF \cite{guo2022neural} performs well, 
but it is limited by the Manhattan-world assumption. Our method with dense depth outperforms Manhattan-SDF \cite{guo2022neural}, verifying the effectiveness of the plane constraints (compared with the Manhattan-world assumption). 


We finally compare our method with two methods using dense depth \textbf{with 3D supervision}. Atlas \cite{murez2020atlas} achieves high completeness and recall due to its TSDF completion capability, but the accuracy and precision are low. NeuRIS \cite{wang2022neuris} performs much well in indoor scenes, but it needs pre-trained networks finetuned by 3D supervision to provide normal maps to supervise the networks. Our method using dense depth without any 3D supervision achieves comparable results with NeuRIS \cite{wang2022neuris} that utilizes pre-training networks.

\section{Conclusion}
In this paper, we present a novel 3D scene reconstruction method without 3D supervision. 
Our method can reconstruct indoor scenes with accessible 2D images as supervision by using sparse depth under the plane constraints.
The sparse depth obtained by using geometry constraints can guarantee the good reconstruction quality of the scene regions with complex geometry structures.
The plane constraints can guarantees the good reconstruction quality of the scene regions with large low-textured planes by making large planes keep parallel or vertical to the wall or floor.
Experiments can show that our method reconstructs the scene completely and accurately, and performs competitive reconstruction quality to the methods that require 3D data or pre-trained networks finetuned by 3D data. 


\bibliographystyle{ieee_fullname}
\bibliography{egbib}

\end{document}